\documentclass[a4paper,twoside]{article}

\usepackage{epsfig}
\usepackage{subcaption}
\usepackage{calc}
\usepackage{amssymb}
\usepackage{amstext}
\usepackage{amsmath}
\usepackage{amsthm}
\usepackage{multicol}
\usepackage{pslatex}
\usepackage{apalike}

\usepackage{pifont}
\usepackage{enumitem}
\usepackage[dvipsnames]{xcolor}

\usepackage{SCITEPRESS}

\author{\authorname{
Steffen Illium\sup{1},
Thore Schillman\sup{1},
Robert Müller\sup{1},
Thomas Gabor \sup{1} and 
Claudia Linnhoff-Popien \sup{1}}
\affiliation{\sup{1}Institute for Informatics, LMU Munich, Oettingenstr. 67, Munich, Germany}
\email{steffen.illium@ifi.lmu.de}
}

\begin{document}

\newlist{todolist}{itemize}{2}
\setlist[todolist]{label=$\square$}
\newcommand{\cmark}{\ding{51}}%
\newcommand{\xmark}{\ding{55}}%
\newcommand{\done}{\rlap{$\square$}{\raisebox{2pt}{\large\hspace{1pt}\cmark}}%
\hspace{-2.5pt}}
\newcommand{\wontfix}{\rlap{$\square$}{\large\hspace{1pt}\xmark}}
\newtheorem{mydef}{Definition}

\title{Empirical Analysis of Limits for Memory Distance\\in Recurrent Neural Networks}

\abstract{
Common to all different kinds of recurrent neural networks (RNNs) is the intention to model relations between data points through time. 
When there is no immediate relationship between subsequent data points (like when the data points are generated at random, e.g.), we show that RNNs are still able to remember a few data points back into the sequence by memorizing them by heart using standard backpropagation. 
However, we also show that for classical RNNs, LSTM and GRU networks the distance of data points between recurrent calls that can be reproduced this way is highly limited (compared to even a loose connection between data points) and subject to various constraints imposed by the type and size of the RNN in question. 
This implies the existence of a hard limit (way below the information-theoretic one) for the distance between related data points within which RNNs are still able to recognize said relation.
}

\onecolumn \maketitle \normalsize \setcounter{footnote}{0} \vfill

\section{\uppercase{Introduction}}
\label{sec:introduction}

Neural networks (NNs) have established themselves as the premier approach to approximate complex functions~\cite{engelbrecht2007computational}. 
Although NNs introduce comparatively many free parameters (i.e., weights that need to be assigned to correctly instantiate the function approximation), they can still be trained efficiently using backpropagation~\cite{rumelhart1986learning}, often succeeding the performance of more traditional methods. 
This makes neural networks rather fit for tasks like supervised learning and reinforcement learning, when hard-coded algorithms are complicated to build or possess long processing times. 
In general, NNs can be trained using backpropagation whenever a gradient can be computed on the loss function (i.e., the internal function is differentiable). 

One of the crucial limitations of standard NNs is that their architecture is highly dependent on rather superficial properties of the input data.
To change the architecture during training, one has to deviate from pure gradient based training methods~\cite{stanley2002evolving}. 
However, these methods still do not allow the shape of input data to change dynamically, thus allowing various input lengths to be processed by the same NN model.

Common use cases like the classification of (various length) time series data have inspired the concept of \emph{Recurrent Neural Networks} (RNNs,~cf.~\cite{rumelhart1986learning}): They apply the network to a stream of data piece by piece, while allowing the network to connect to the state it had when processing the previous piece. 
Thus, information can be carried over when processing the whole input stream. 
We provide more details on the inner mechanics of RNNs and their practical implementation in Section~\ref{sec:foundations}. 
\emph{Long Short-Term Memory networks} (LSTMs,~cf.~\cite{hochreiter1997long}) are an extension of RNNs: They introduce more connections (and thus also more free parameters) that can handle the passing of state information from input piece to input piece more explicitly (again, cf. Section~\ref{sec:foundations} for a more detailed description). 
Based on the LSTM architecture, Cho et al. introduced the \emph{Gated Recurrent Unit} (GRU) with smaller numbers of free parameters but comparable training achievements depending on the task~\cite{cho2014learning,chung2014empirical}.

All kinds of RNNs have been built around the idea that carrying over some information throughout processing the input data stream allows the networks to recognize connections and correlations between data points along the stream. 
For example, when we want to predict the weather by the hour, it might be beneficial to carry over data from the last couple of days as to know how the weather was at the same time of day yesterday. 
However, it is also clear that (as the number of connections "to the past" is still fixed given a single instance of an RNN) the amount of information that can be carried over is subject to a hard theoretical limit. 
In practice, of course, NNs are hardly capable of precisely manipulating every single bit of their connections, so the practical limit for remembering past pieces of the input stream must be way below the inputs vector information memory size.

In this paper, we first show that the memorization of random inputs can in fact be learned by RNNs and then recognize a hard practical limit to how far a given RNN architecture can remember the past.
We construct the memorization experiment in Section~\ref{sec:methodology}). 
Our experiments show that a distance limit, i.e., the loss of the ability to store and reproduce values, occurs very prominently, of course varying with the type and architecture of the RNN~(cf. Section~\ref{sec:results}).

While these experiments explore the basic workings of RNNs, we are not aware of other research assessing a comparable limitation on RNNs. Closest related work is discussed in Section~\ref{sec:related}. 
The existence of a hard limit in memory distance may have several implications on the design and usage of RNNs in theory and practice like how big an RNN needs to be so that it is able to remember the weather from 24 hours ago, e.g., which we briefly discuss as this paper concludes (cf. Section~\ref{sec:conclusion}).

\section{\uppercase{Foundations}}
\label{sec:foundations}

    \begin{figure*}[t]
    	\begin{minipage}[t]{.48\textwidth}
    		\includegraphics[width=.9\linewidth]{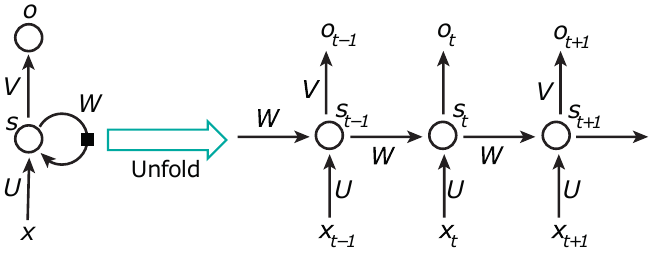}
    		\caption{"A recurrent neural network and the unfolding in time of the computation involved in its forward computation." Image taken from \cite{lecun2015deep}.}
    		\label{fig:rnn_unfold}
    	\end{minipage}%
    	\hfill
    	\begin{minipage}[t]{.48\textwidth}
    		\includegraphics[width=.9\linewidth]{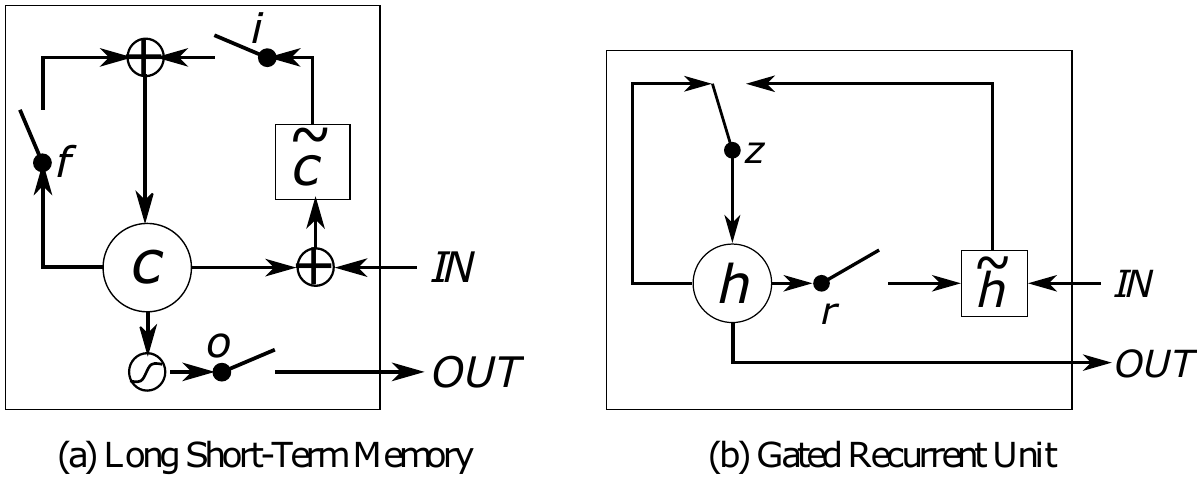}
    		\caption{"Illustration of (\textbf{a}) LSTM and (\textbf{b}) gated recurrent units. (\textbf{a}) $i$, $f$ and $o$ are the input, forget and output gates, respectively. $c$ and $\tilde{c}$ denote the memory cell and the new memory cell content. (\textbf{b}) $r$ and $z$ are the reset and update gates, and $h$ and $\tilde{h}$ are the activation and the candidate activation." Image taken from \cite{chung2014empirical}.}
    		\label{fig:lstm_gru_cell}
    	\end{minipage}%
    \end{figure*}

We consider the three most popular types of RNNs for experiments on their respective memory limits.
In this section, we give a quick recap on their most relevant properties.

\subsection{Recurrent Neural Networks}
\emph{Recurrent Neural Networks} (RNNs) were the first models introduced to process data sets that imply a temporal dependency on sequential inputs~\cite{schmidhuber2015deep}.

While (given an input vector $\mathbf{x} = \langle x_i \rangle_{1 \leq i \leq q, i \in \mathbb{N}}$ of length $q \in \mathbb{N}$) one element $x_i$ at a time is processed, a hidden state vector $\mathbf{h}$ carries the history of all the past elements of the sequence~\cite{lecun2015deep}.
Finally, a $\textit{tanh}$ activation function shifts the internal states to the nodes of the output layer.
Given a sequential input $\mathbf{x}$ and an activation function $g$, the output state for every time step $t$ is $h_t = g(Wx_t + Uh_{t-1})$~\cite{chung2014empirical}.
RNNs can be thought of as single layer of neural cells, which is reused multiple times on a sequence while taking track of all subsequent computations (cp. Figure~\ref{fig:rnn_unfold}). Therefore, every position of the output vector $z$ is influenced by the combination of every previously evaluated input as described by:

\begin{equation}
	z = tanh((x \odot wx + b) + (ws \odot s))
	\label{eq:tanh}
\end{equation}

\subsection{Long Short-Term Memory Networks}
While regular RNNs are known for their problems in understanding long-time dependencies as the gradient either explodes or vanishes \cite{lecun2015deep,chung2014empirical,schmidhuber2015deep}, Hochreiter and Schmidhuber introduced the~\emph{Long Short-Term Memory~(LSTM)} unit~\cite{hochreiter1997long}.
The main difference lies in the "connection to itself at the next time step through a memory cell"~\cite{lecun2015deep} as well as a forget gate, that regulates the temporal connection (cp. Figure~\ref{fig:lstm_gru_cell}, a).
Layers equipped with LSTM cells can learn what to forget while remembering valuable inputs in long temporal sequences.
However, this functionality comes at a high computational cost in total trainable parameters $N = 4 \cdot (mn + n^2 + n)$ where $m$ is the input dimension, and $n$ is the cell dimension.
This constitutes a four-fold increase in comparison to a RNN~\cite{lu2017simplified}.


\subsection{Gated Recurrent Unit Networks}

Cho et al. proposed a new LSTM variant, \textit{Gated Recurrent Unit (GRU)} with reduced parameters and no memory cell.
At first "the reset gate $r$ decides whether the previous hidden state is ignored'', then "the update gate $z$ selects whether the hidden state is to be updated''. 
In combination with the exposure of the hidden state, the internal decision gates could be reduced from four to "only two gating units'', so that the overall number of trainable parameters per layer activation decreases (see~\cite{cho2014learning,chung2014empirical}; cp. Figure~\ref{fig:lstm_gru_cell}, a). 
In short, a GRU cell is a hidden layer unit that "includes a reset gate and an update gate that adaptively control how much each hidden unit remembers or forgets while reading/generating a sequence''~\cite{cho2014learning}. 
Even though \cite{chung2014empirical} did not prefer GRU over LSTM in an empirical testing setup, they saw that the usage "may depend heavily on the dataset and corresponding task''.

\subsection{Echo State Networks}

\begin{figure}[ht]
	\centering
	\includegraphics[width=0.4\textwidth]{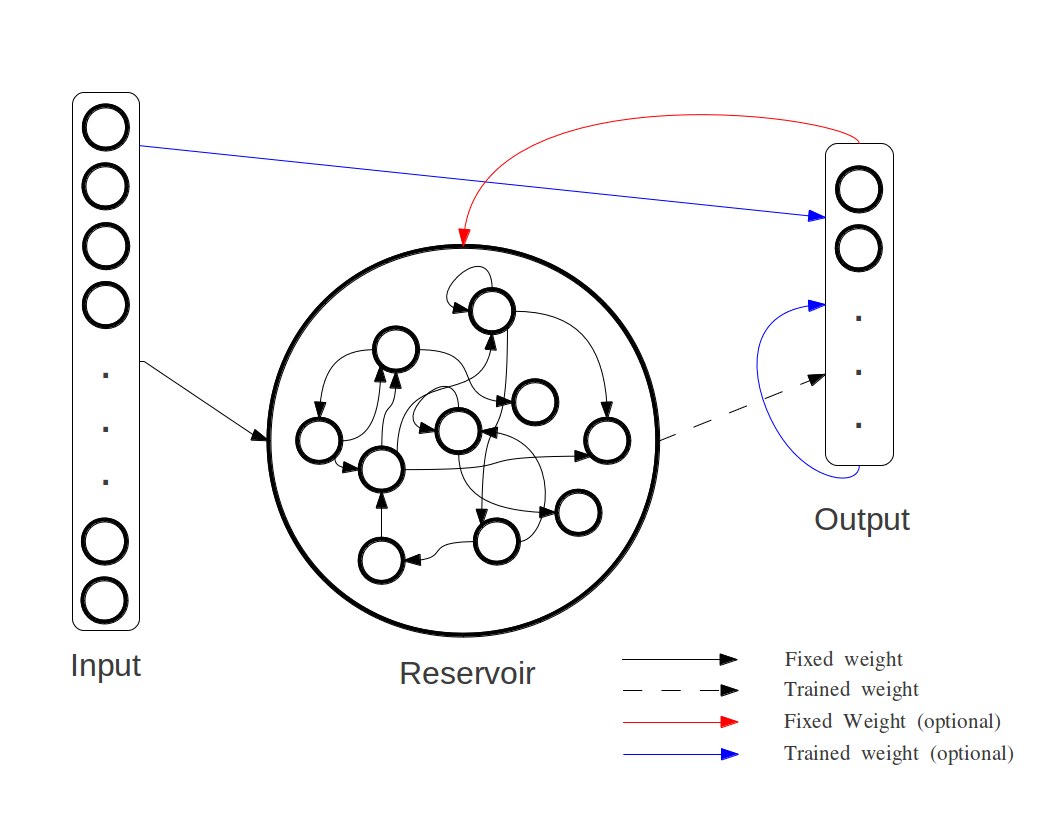}
	\caption{An ESN with all possible kinds of connections. The legend denotes which ones are fixed, trainable and optional.}
	\label{fig:esn}
\end{figure}

\emph{Echo state networks} (ESNs) are a type of \emph{reservoir computer}, independently developed by Herbert Jaeger~\cite{jaeger2001echo} and Wolfgang Maass~\cite{maass2002real}. Reservoir computing is a framework which generalises many neural network architectures utilising recurrent units. ESNs differ from "vanilla" RNNs in that the neurons are sparsely connected (typically around 1\% connectivity) and the reservoir weights are fixed. Also, connections from the input neurons to the output as well as looping connections in the reservoir (hidden layer) are allowed, see Figure~\ref{fig:esn}.

Unlike most recurrent architectures in use today, ESNs are not trained via \emph{backpropagation though time} (BPTT), but simple linear regression on the weights to the output units. Having fixed reservoir weights makes the initialization even more important. Well initialized ESNs display the \emph{echo state property}, which guarantees, after an initial transient, that the activation state is a function of the input history:
\begin{equation}
	h_t = E(...,x_{t-1}, x_t)
\end{equation}
Here, $E$ denotes an \textit{echo function} $E=(e_1,...,e_{N})$ with $e_i:U^\mathbb{-N}\rightarrow\mathbb{R}$ being the echo function of the $i$-th unit.

Several conditions for echo states have been proposed, the most well-known stating that the spectral radius and norm of the reservoir matrix must be below unity~\cite{jaeger2001echo}. This has been shown to neither be sufficient nor necessary~\cite{yildiz2012re,mayer2017echo}, but remains a good rule of thumb.

The amount of information from past inputs that can be stored in the ESNs transient state dynamics is often called its short-term memory and is typically measured by the \emph{memory capacity} (MC), proposed by Jaeger~\cite{jaeger2001echo}. It is defined by the coefficient of correlation between input and output, summed over all time delays
\begin{equation}
    MC = \sum_{k=1}^{\infty}MC_{k} = \sum_{k=1}^{\infty}\frac{cov^2(x(t-k), y_k(t))}{Var(x(t))*Var(y_k(t))}
\end{equation}
and thus, measures how much of the inputs variance can be recovered by optimally trained output units.
It has been shown by Jaeger that the MC of an ESN with linear output units is bounded above by its number of reservoir neurons N. 

ESNs predate the deep learning revolution. Nonetheless, deep variants of ESNs have recently been proposed \cite{gallicchio2017deep} to make use of the advantages of layering. Reservoir computing is an active field of research, as the models are less expensive to train, often biologically plausible~\cite{buonomano1995temporal} and allow for precise description of state dynamic properties due to the fixed weights of the reservoir.

\section{\uppercase{Methodology}}
\label{sec:methodology}

To reveal the limits of distance in which information can be carried over from past states $h_{t-n}$ through recurrent connections of RNN type networks, we construct a task in which the RNN is forced to remember a specific input of the input data stream $\mathbf{x}$.

This implies that there must not be any correlation between the pieces of the input data stream so that there is no way to reconstruct past pieces without fully passing in their values through the recurrent connections to all future applications of the RNN.
We implemented this experiment using random-length sequences of random numbers as input data to the RNN.
The task (cf. Def.~\ref{def:mem}) is the reproduction of the $p$th-from-last input throughout a training setup for a set $p \in \mathbb{N}$.

However, since the sequences are of random lengths, the RNN cannot recognize which piece of the input stream will be needed until the very last application.
This forces the RNN to carry over all the information it can from the input stream.
We argue that this is the canonical experiment to test memorization abilities of RNNs.
This task definitions are in stark contrast to common applications of RNNs where one explicitly expects some correlations between the pieces of the input sequence and thus uses RNNs to approximate a model of said connections.

\begin{mydef}[Random Memorization Task]
\label{def:mem}
	Let $q$ be a random number uniformly sampled from $[10;15] \subseteq \mathbb{N}$.
	Let $\mathbf{x} = \langle{x_i}\rangle_{1 \leq i \leq q, i \in \mathbb{N}}$ be a sequence of random numbers $x_i$ each uniformly sampled from $[0;1] \subseteq \mathbb{R}$.
    Let $\mathcal{R}$ be an RNN that is applied to the sequence $\mathbf{x}$ in order with the result $\mathcal{R}(\mathbf{x})$ of the last application.
	The \emph{memorization task of the $p$th-from-last position} is given by the minimizing goal function $f_p(\mathcal{R}, \mathbf{x}) = | x_{q+1-p} - \mathcal{R}(\mathbf{x}) |$.
\end{mydef}

Note that the random length of the sequence is substantial.
Otherwise, an RNN trained to reproduce the $p$th-from-last position may count the position of the pieces of the input sequence from the start, making it rather easy to remember the $q$th-from-last (i.e., first) position.
In complexity terms, this task requires space complexity of O(1).
These claims are supported by our experiments in Section~\ref{sec:results}, where we also test a fixed-length memorization task:

\begin{mydef}[Fixed-Length Random Memorization Task]
\label{def:mem-fixed}
	Let $q=10$. For all else, refer to Def.~\ref{def:mem}.
\end{mydef}

\section{\uppercase{Experimental Results}}
\label{sec:results}

To evaluate the task described in Definitions~\ref{def:mem} we conducted several experiments with different configurations of RNNs, LSTMs, and GRUs.
A configuration is specified by the number of layers $l$ of the network and the number of cells per layer $c$.
We disregarded any complex architectural pattern by just examining standard fully connected recurrent layers as well as modern network training techniques such as gradient clipping or weight decay in favor of comparability.
Layers' outputs from more than $c\geq1$ cells are added up.
Weight updates are applied by \emph{stochastic gradient descent} (SGD) and the \textit{truncated backpropagation} algorithm. 
Training data is generated by sampling from a Mersenne Twister pseudo-random number generator.
Every network configuration was trained for up to 5000 epochs for 5 different random seeds if not stated otherwise.

\begin{figure}[t]
	\centering
		\includegraphics[width=.99\columnwidth]{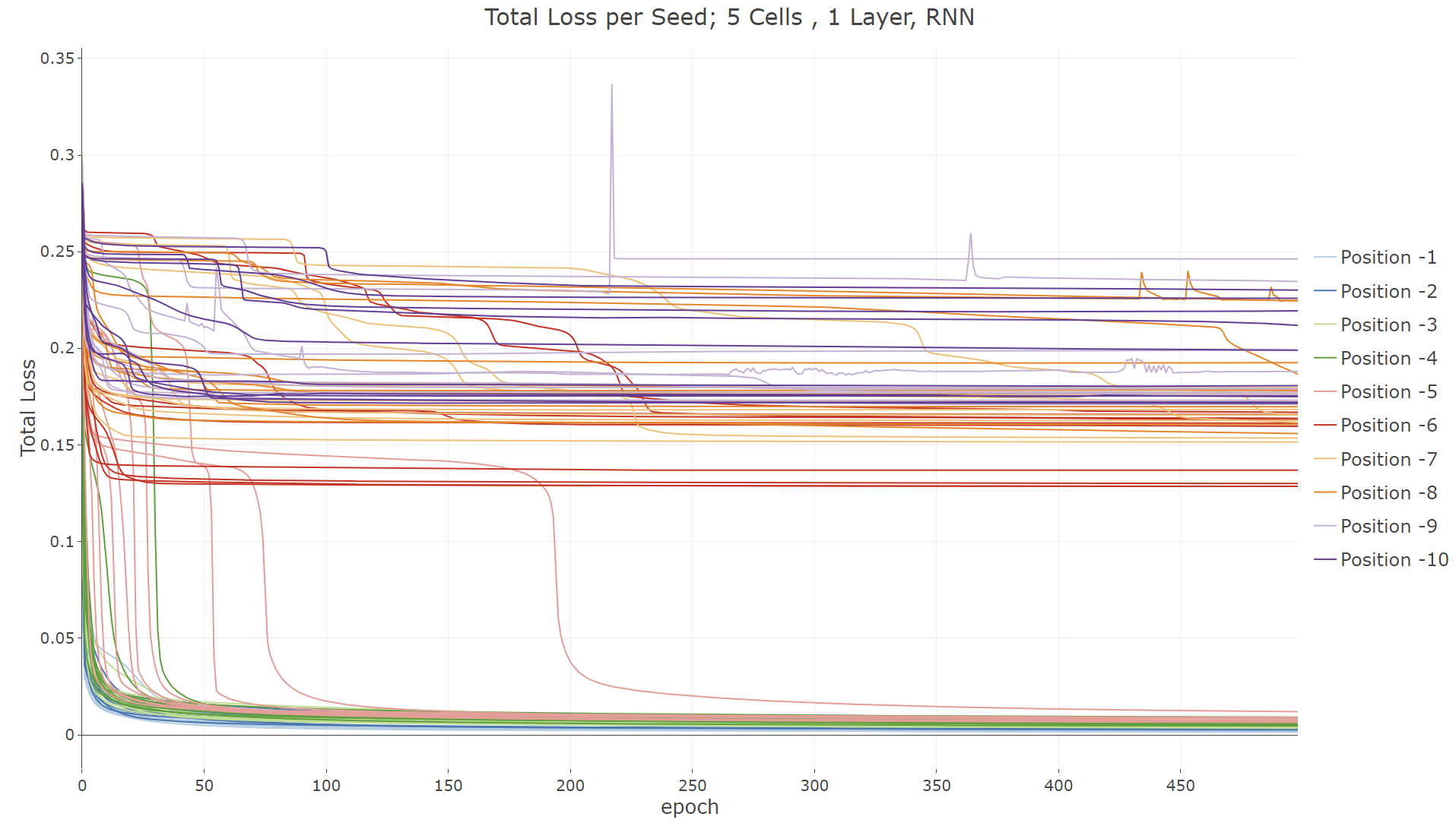}
		\caption{Absolute loss for an \textbf{RNN} with $l = 1$ layer and $c = 5$ cells when asked to reproduce the $p$th-from-last position (for $p \in [1;10]$) of a random sequence of random length $q \in [10;15]$. Every color shows $10$ different training runs (with different random seed) for a single $p$.}
		\label{fig:RNN-lines}
\end{figure}

\textbf{Random Memorization}~We start by evaluating the random memorization task (cf. Definition~\ref{def:mem}).
We run a simple scenario with $l = 1$ layer and $c = 5$ cells within that layer. 
We conducted $10$ independent training runs each for every position to memorize from "position -1", i.e., the last position or $p=1$, to "position -10", i.e., the $10$th-from-last position or $p=10$.
Figure~\ref{fig:RNN-lines} shows the results for an RNN plotting the loss over training time: Reproducing positions $1 \leq p \leq 4$ is learned quickly so that most runs overlap at the bottom of the plot.
$p = 5$ seems to be harder to learn as some runs take distinctively more time to reach a loss close to zero, but the $5$th-from-last random memorization task is still learned reliably. Positions $6 \leq p \leq 8$ seem to stagnate not being able to reduce the loss beyond around $0.15$, which still means that some information about these positions' original value is being passed on, though.
However, the network greatly loses accuracy and cannot reliably reproduce the values. We observe another case in most runs of positions $p = 9$ and $p = 10$, where the loss seems to stagnate at around $0.25$: Note that for the random memorization task of sequence values $x_i \in [0;1] \subseteq \mathbb{R}$ a static network without any memorization can simply always predict the fixed value $\mathcal{R}(\mathbf{x}) = 0.5$ to achieve an average loss of $0.25$ on a sequence of random numbers.
At least, that trick is easily learned by all setups of RNNs.

\begin{figure}[t]
    \centering
		\includegraphics[width=.99\columnwidth]{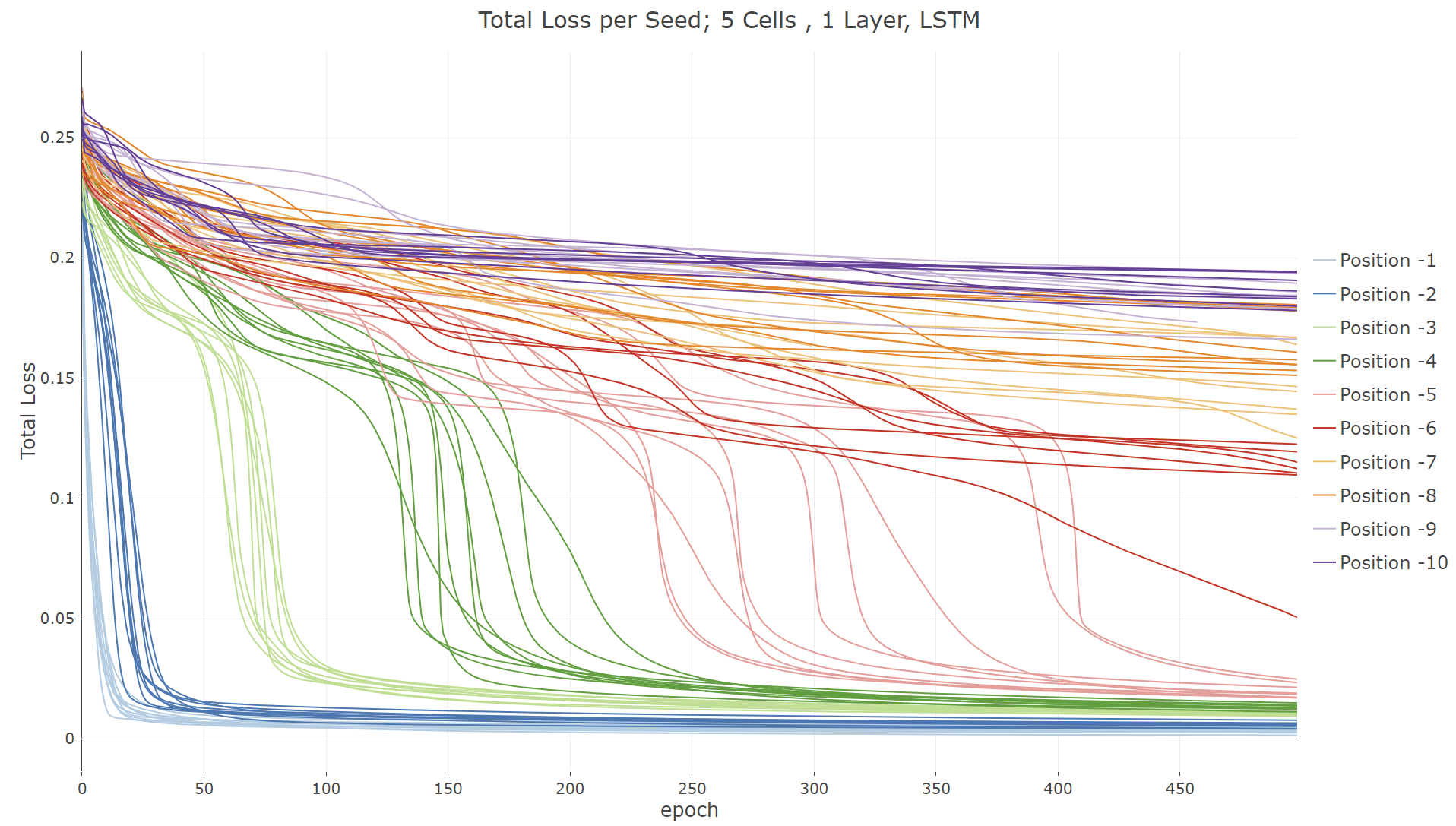}
		\caption{Absolute loss for an \textbf{LSTM} with $l = 1$ layer and $c = 5$ cells when asked to reproduce the $p$th-from-last position (for $p \in [1;10]$) of a random sequence of random length $q \in [10;15]$. Every color shows $10$ different training runs (with different random seed) for a single $p$.}
		\label{fig:LSTM-lines}
\end{figure}

We compare those results to experiments executed for LSTMs in Figure~\ref{fig:LSTM-lines}. First, we notice that the LSTM takes much longer training time for the simpler positions $1 \leq p \leq 5$, which are all learned eventually but show a clear separation. This result is intuitive considering the additional number of weights (total size $4 \cdot (mn+n^2+n)$ for $m$ input elements and $n$ neurons), i.e., trainable parameters, for the same size of network compared to RNNs (total size $mn+n^2+n$). On the other hand, we can observe that LSTMs still show some learning progress on all positions $p$ after the 500 epochs plotted here, especially having a run of position $p=6$ taking off towards near-optimal loss. Thus, there is reason to believe that LSTMs may eventually prove to be considerably more powerful than RNNs given significantly longer training times on our task. 


\begin{figure*}[htb]
	\centering
	\includegraphics[width=\textwidth]{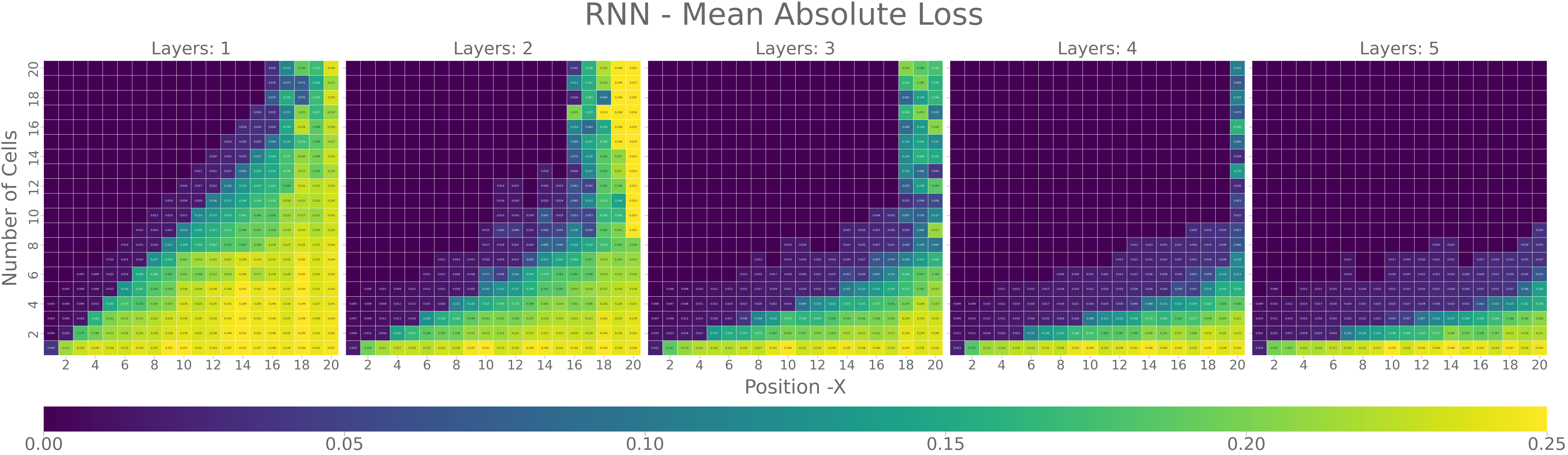}
	\caption{Mean absolute loss for various configurations of \textbf{RNNs} with $l \in [1;5]$ layers with $c \in [1;20]$ cells each when asked to reproduce the $p$th-from-last position (for $p \in [1;20]$) of a random sequence of random length $q \in [10;15]$. Every data point is averaged over $5$ runs.}
	\label{fig:RNN}
\end{figure*}

For a more complete overview of the limits of performance in the random memorization task we ran this experiment for different number of layers $l \in [1;5]$ and cells per layer $c \in [1;20] \subseteq \mathbb{N}$ as well as for the $p$th-from-last positions $p \in [1;20] \subseteq \mathbb{N}$ as a target. Figure~\ref{fig:RNN} shows the results for RNNs. The color of the respective dots encoding the average loss over $5$ independent runs, the scale of the loss is again fitted to $0.25$ as all RNNs manage to learn that fixed guess as discussed above. We can observe that the $1$-layer $1$-cell network at the bottom left only learns to reproduce the last value of the random sequence, which it can do without any recurrence by just learning the identity function. 
It is thus clear why all RNNs manage to solve this instance. 
Furthermore, we can clearly see a monotone increase in difficulty when increasing the position, i.e., forcing the network to remember more of the sequence's past. This shows that no network can learn a generalization of the task that can memorize arbitrary positions as presumed when constructing the task.

\begin{figure*}[htb]
	\centering
	\includegraphics[width=\textwidth]{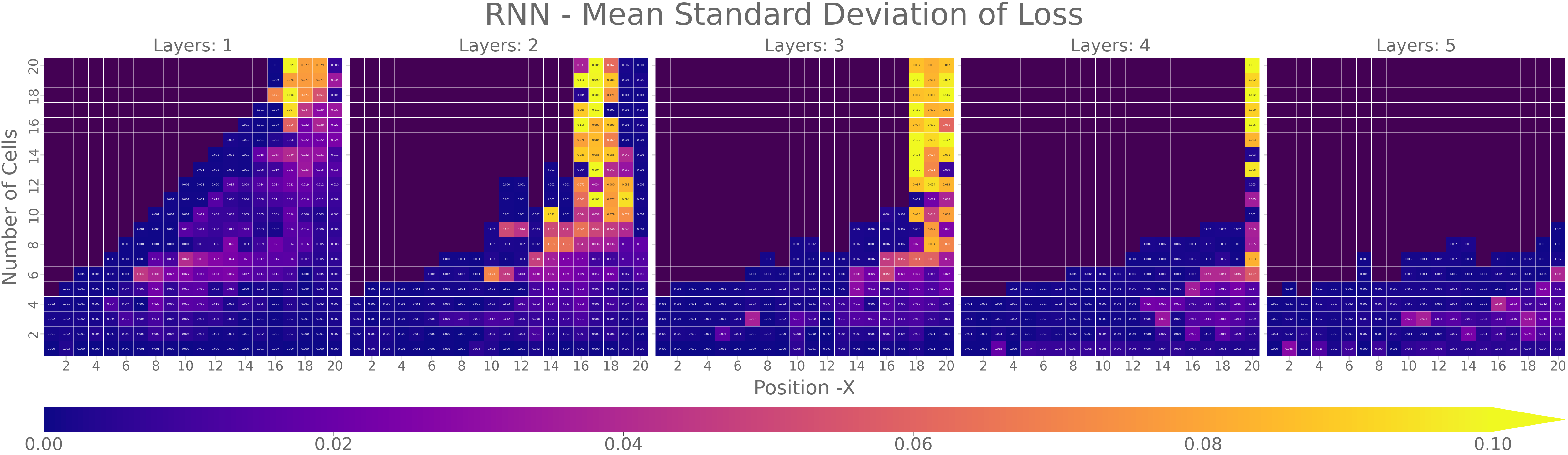}
	\caption{Standard Deviation for various configurations of \textbf{RNNs} with $l \in [1;5]$ layers with $c \in [1;20]$ cells each when asked to reproduce the $p$th-from-last position (for $p \in [1;20]$) of a random sequence of random length $q \in [10;15]$. Every data point is averaged over $5$ runs.}
	\label{fig:RNN_STD}
\end{figure*}

What is possibly the most interesting result of these experiments is that the number of instances resulting in average losses of around $[0.05; 0.2]$ (colored in light blue, green) is relatively small: Certain positions can either be reproduced very clearly or barely better than the fixed guess (purple, orange), cf.~Figure~\ref{fig:RNN_STD}. This implies a strong connection between the amount of information that can be passed on throughout the RNN. Furthermore, we can observe that more cells and more layers (which also bring more cells to the table) clearly help in solving the memorization task. Interestingly, even parallel architectures as shown for the networks with only one layer ($l = 1$) become better with more cells. The connection here appears almost linear: Within our result data, a one-layer network with $c=n$ cells can always memorize the $n$th-from-last position ($p=n$) very well.

\begin{figure*}[tb]
	\begin{minipage}[t]{.48\textwidth}
		\includegraphics[width=.9\textwidth]{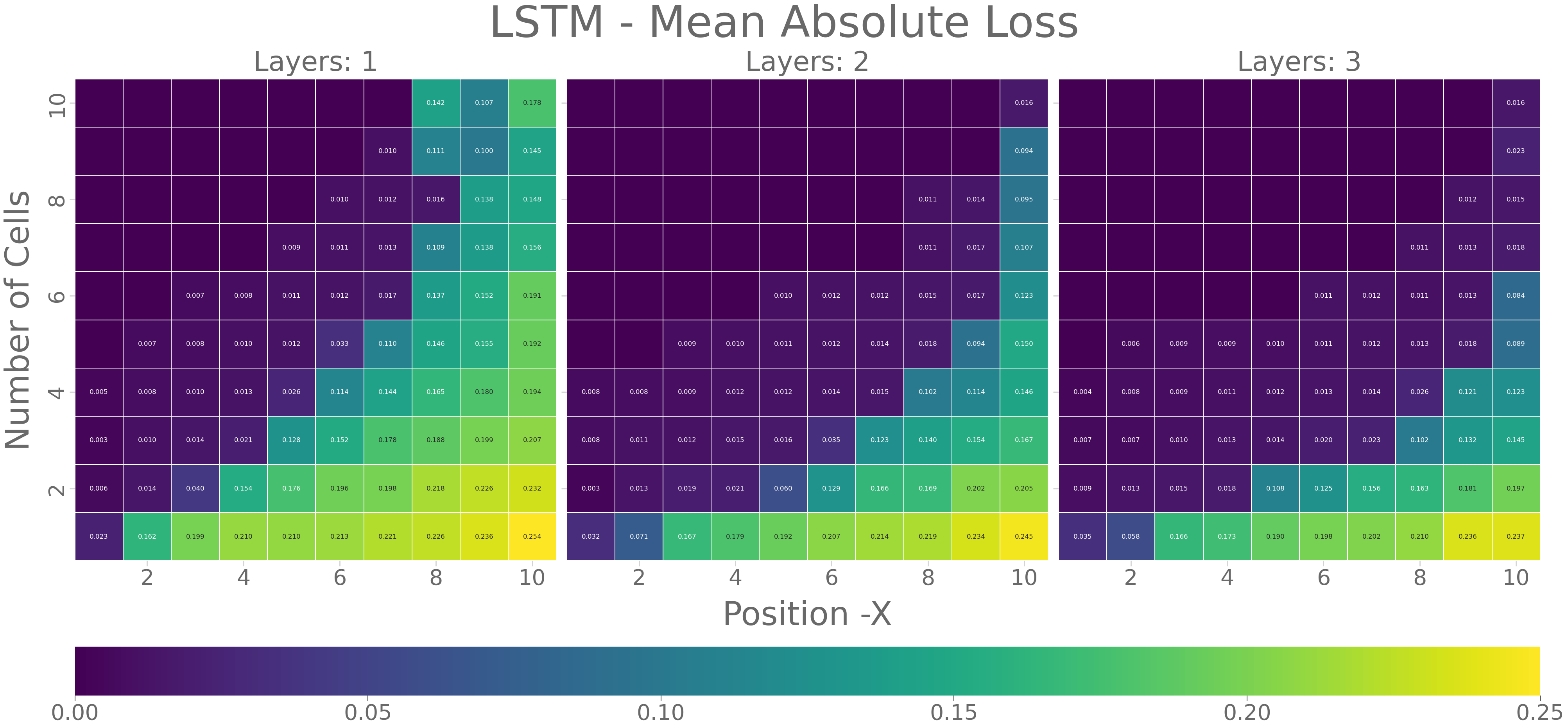}
	\end{minipage}%
	\hfill
	\begin{minipage}[t]{.48\textwidth}
		\includegraphics[width=.9\textwidth]{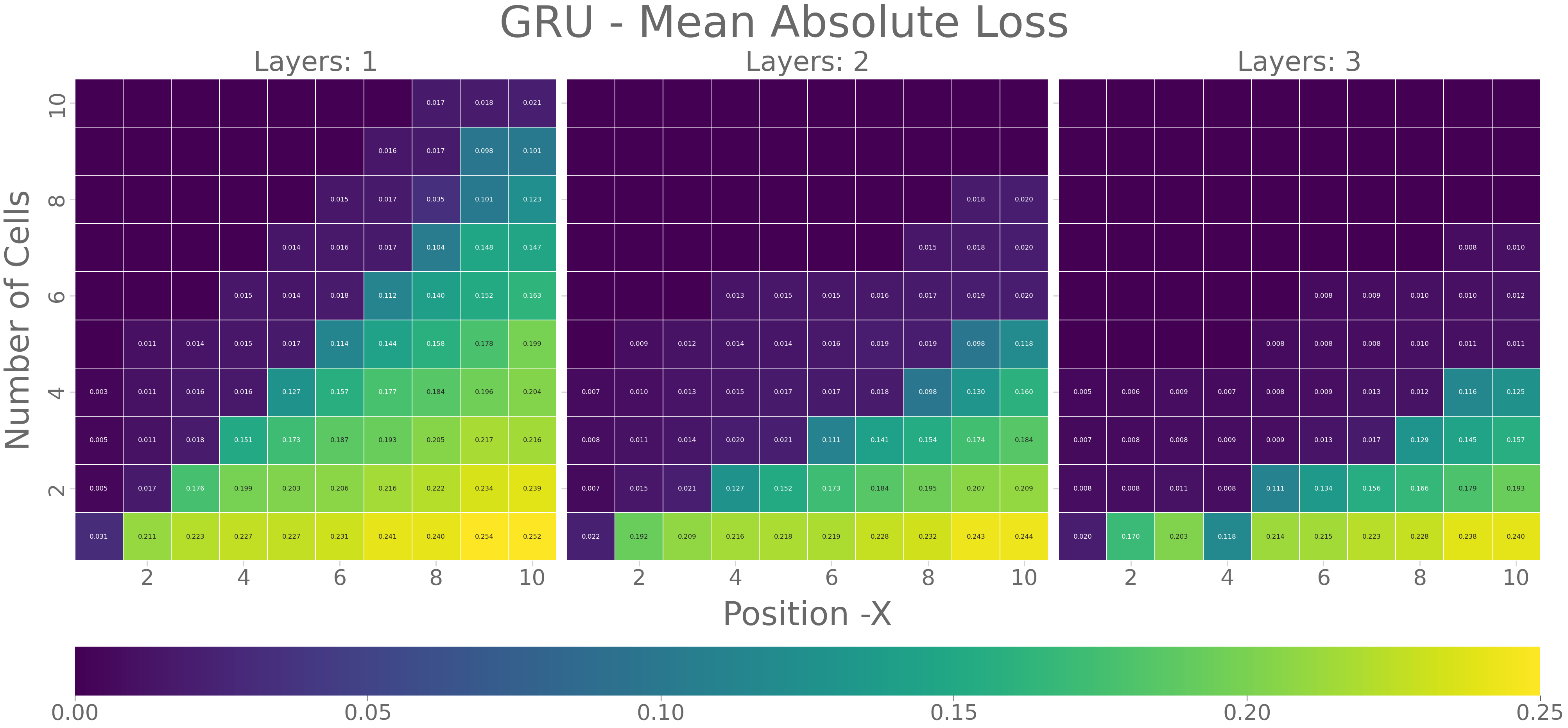}
	\end{minipage}%
	\caption{Mean absolute loss for various configurations of \textbf{LSTMs}~(left) and \textbf{GRU}~(right) with $l \in \{1,2,3\}$ layers with $c \in [1;10]$ cells each when asked to reproduce the $p$th-from-last position (for $p \in [1;10]$) of a random sequence of random length $q \in [10;15]$. Every data point is averaged over $10$ runs.}
	\label{fig:LSTM_GRU}
\end{figure*}

Figure~\ref{fig:LSTM_GRU} shows the same experimental setup as described above for LSTM networks.
In line with the observation about the loss plots (cf. Figure~\ref{fig:LSTM-lines}), LSTMs do not perform as well as RNNs on this task, i.e., not satisfyingly solving any position $p=10$ instance ($\textit{mean\_absolute\_loss} \geq 0.04$).
However, as described above we conjecture this stems from the fact that the higher number of weights present in an LSTM network requires substantially longer training times beyond what could be performed as part of this experiment.
These results still have practical implications: As this relatively small task consumes large computational resources, LSTMs seem to be inefficient at least at (smaller instances of) the random memorization task, at least for single layer networks.

We also conducted the same experiments for GRU networks, as shown in Figure~\ref{fig:LSTM_GRU}. The GRUs' behavior appears most similar to LSTMs'.
We will use this experiment as a basis to refrain from showing GRU plots for other experiments since GRUs have performed that similarly throughout all our experiments.
However, note that GRUs require fewer parameters ($N = 3\cdot(mn+n^2+n)$), making them interesting for efficient practical applications~\cite{dey2017gate}.

Our results indicate that Jaegers upper bound on ESN short-term memory remains valid in the context of deep RNN architectures trained by backpropagation (RNN, LSTM, GRU).
For stacked, multi-layer architectures, the maximum memory capacity could not be reached. Instead, we observed an upper bound of $MC \leq N-(l-1)$ with $N=c*l$. 

\subsection{Fixed-Length Random Memorization}

\begin{figure}[t]
	\centering
	\includegraphics[width=.5\textwidth]{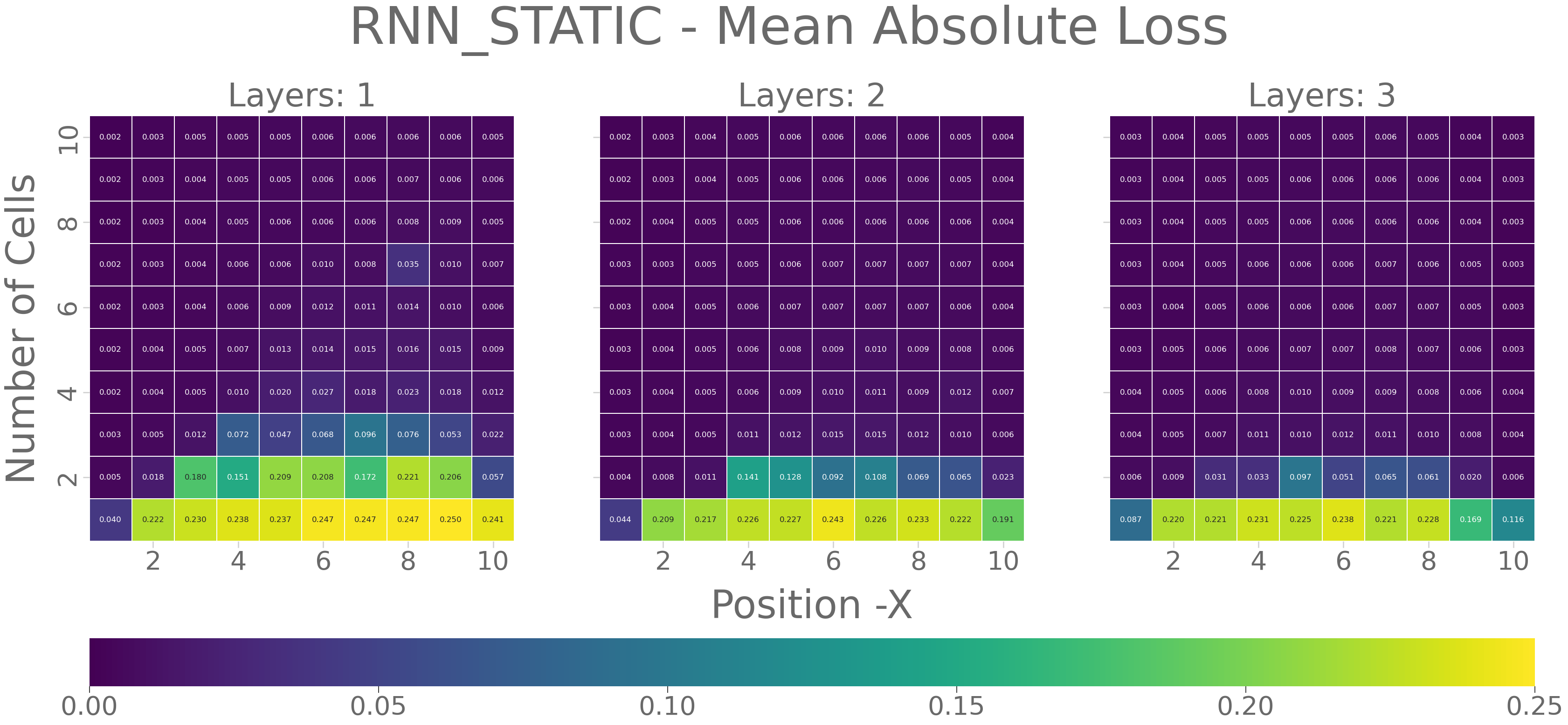}
	\caption{Mean absolute loss for various configurations of \textbf{RNNs} with $l \in \{1,2,3\} $ layers with $c \in [1;10]$ cells each when asked to reproduce the $p$th-from-last position (for $p \in [1;10]$) of a random sequence of fixed length $q = 10$. Every data point is averaged over $10$ runs.}
	\label{fig:RNN-static}
\end{figure}

We now quickly discuss the scenario of Definition~\ref{def:mem-fixed}: In this case, we fix the length of the random sequence $q = 10$. This means that in theory it is possible to "count" the positions from the start and thus know exactly when the currently processed piece of the input data stream needs to be memorized, sparing the effort of memorizing as many of them as possible until the end. We thus consider this task strictly easier and our experimental results shown in Figure~\ref{fig:RNN-static} support this. However, the strategy of "counting from the start" seems to be subject to some memory limitations as well. Thus, while every instance that is easily learned in the case of variable-length sequences is also learnable in this scenario, further learnable instances appear from the right of the plot, i.e., $p = 10$ for the 1-layer 2-cells network and $p \geq 8$ for the 2-layers 2-cells-each network. This shows that the RNN is not exactly able to count through the sequence it is given but able to memorize pieces from the start, within a similar limit as from the end of the sequence. As LSTMs and GRUs show a similar behavior, we omit the respective plots for brevity.

\section{\uppercase{Related Work}}
\label{sec:related}

Recent studies on recurrent neural networks mostly focused on memory afforded by attractor dynamics or weight changes while learning. Both are only useful in the context of temporally correlated data, as the model settling in a low dimensional state space would not be desirable when every data point is equally important. As RNNs have been conceived precisely to recognize and model dependencies in sequential data, their ability to deal with i.i.d. or uncorrelated input might seem to be of little importance. We argue that this capability is an integral part of RNN memory and performance, which has long been recognized in the reservoir computing literature. For tasks where the output should depend strongly on the whole input history, we require a unique, clearly distinguishable hidden state for every input. Using uncorrelated data, we can approximate the total amount of information the architecture can store in its transient activation dynamics. 

In addition, and specific to RNN-architectures a lot of empirical evaluations on various LSTM and RNN configurations and tasks already exist.
Hochreiter and Schmidhuber for example evaluated LSTM cells from the beginning. 
They conjectured that gradient-based approaches suffer from practical inability to precisely count discrete time steps and therefore assume the need for an additional counting mechanism~\cite{hochreiter1997long}.
However, our experiments show that an explicit counting mechanism might not be necessary, even for models, trained on random sequences (i.e., pure noise).
At least for the extend of our observations.


First experiments on the ability of an RNN to carry information over some distance in the input sequence were conducted~\cite{cleeremans1989finite}. 
Likening an RNN to a finite state automaton (for more details on that line of thought~\cite{giles1992learning}), they trained a network to recognise an arbitrary but fixed regular grammar. 
Interestingly, they concluded that a recurrent network can encode information about long-distance contingencies only as long as information about critical past events is relevant at each time step, which again contrasts our results on random sequences.

Little research has been performed on random sequences of variable length as inputs for RNNs. One could argue that experiments~4 and~5 by Hochreiter and Schmidhuber are of relevance here~\cite{hochreiter1997long}.
However, we like to stress that our work is focused on random sequences both in length and value to reproduce an element $x_i$ on a specific position without providing any form of hint such as additional labelling or encoding.
We see our work in contrast to these early experiments, as we test for maximal signal remoteness for a given computational budget and not necessary for memory endurance in general.
In other words,  given an input vector of randomly generated length, they ask for the binary answer to whether a specific symbol came before another. This tests if the gradient can uphold for enough recurrent steps so that the two bits of information can be recognized. 
In contrast, given a random input vector of randomly generated length, we ask the networks to reproduce the $p$th-last entry.
We also considered classic RNNs that do not possess gated memory vectors.

Most recently Gonon et al. analysed the memory and forecasting capacities of linear and nonlinear RNN (for independent and non-independent inputs). They stated an upper bound for both while generalizing Jaegers statements regarding MC estimates.~\cite{gonon2020memory}

Additionally, Bengio et al. performed related experiments to investigate the problem of vanishing and exploding gradients. 
Furthermore, their findings explain how robust the transport of information is under the influence of noisy and unrelated information. However, they did not consider pure memorization "by heart''.~\cite{bengio1994learning}

As one might confuse the benchmark problems in this paper with work on the topic of generalization and over-fitting, we would like to point out, that our targeted notion of capacity is very different.
Although unrelated in methodology (they used regular fully connected feed-forward networks), Zhang et al. provide an analysis quite similar in wording: They consider a network's capacity as the amount of input-output pairs from the training set that can be exactly reproduced by the network (after training)~\cite{zhang2016understanding}.
This kind capacity is measured on a whole data set of input vectors.
We consider memory capacity within a single call of a trained network for a single input vector that is passed in piece by piece, which is an entirely different setup.

Neural Turing Machines currently are a promising extension of classical NNs designed specifically to offer more powerful operations to manage a hidden state~(\cite{graves2014neural}). We reckon that the Random Memorization Task is trivial to them.
The same applies for neural networks architectures that involve attention mechanism, like the very prominent Transformer \cite{vaswani2017attention}.

\section{\uppercase{Conclusion}}
\label{sec:conclusion}

We defined the Random Memorization Task for RNNs.
Despite it contradicting the intuition behind the usage of recurrent neural networks, classical RNNs, LSTM and GRU networks were able to memorize a random input
sequence to a certain extent, depending on configuration and architecture.
There is a discernible borderline between past positions in the sequence that could be memorized and those that could not, which correlates with the MC formulated by Jaeger.
We therefor conclude, that Jaegers MC formula is applicable for calculating the memory limit in respect to the RNN's type and architecture.

While our experiments are very limited in scale (due to the already computationally expensive nature of
the experiments), we observe the trend that more cells increase the memory limit. 
The limiting factor that we observed, at least for vanilla-RNN's, is the VEGP.
However, it is important to note that for current RNN applications the ratio of input sequence length and cells inside the RNN is usually in favor of the number of cells, so that we would not expect the memory limit to play an important role in the application of the typical RNN. 
Still, we hope this research represents an additional step towards a theoretical framework concerned with the learnability of problems using specific machine learning techniques.

\bibliographystyle{apalike}
{\small
\bibliography{index.bib}}



\end{document}